\documentclass{article}
\usepackage{preprint}
\usepackage[twoside,letterpaper,includeheadfoot,%
        layoutsize={8.125in,10.875in},%
                layouthoffset=0.1875in,%
                layoutvoffset=0.0625in,%
                left=1.25in,%
                right=1.25in,%
                top=45pt,%
                bottom=45pt,%
                headheight=0pt,% No Header
                headsep=10pt,%
                footskip=25pt]{geometry}
\usepackage[numbers, compress]{natbib}
\usepackage{chngcntr}
\usepackage{graphicx,amsmath,amsfonts,amssymb}
\usepackage{booktabs}
\usepackage[utf8]{inputenc} % allow utf-8 input
\usepackage[T1]{fontenc}    % use 8-bit T1 fonts
\usepackage{hyperref}       % hyperlinks
\usepackage{microtype}      % microtypography
\usepackage{graphicx}
\usepackage[font=footnotesize]{caption}
\usepackage{multirow}
\usepackage{xargs}
\usepackage{ulem}
\usepackage[colorinlistoftodos,prependcaption,textsize=scriptsize]{todonotes}
\newcommandx{\fixme}[2][1=]{\todo[linecolor=red,backgroundcolor=red!25,bordercolor=red,#1]{#2}}
\newcommandx{\changeme}[2][1=]{\todo[linecolor=blue,backgroundcolor=blue!25,bordercolor=blue,#1]{#2}}
\newcommandx{\info}[2][1=]{\todo[linecolor=OliveGreen,backgroundcolor=OliveGreen!25,bordercolor=OliveGreen,#1]{#2}}
\newcommandx{\ablation}[2][1=]{\todo[linecolor=BurntOrange,backgroundcolor=BurntOrange!25,bordercolor=BurntOrange,#1]{#2}}
\newcommandx{\todone}[2][1=]{\todo[linecolor=Gray,backgroundcolor=Gray!25,bordercolor=Gray,#1]{\sout{#2}}}

\title{\LARGE \textbf{H-ARC: A Robust Estimate of Human Performance on the Abstraction and Reasoning Corpus Benchmark} }
\date{}
\author{ \large \textbf{Solim LeGris$^1$}, \textbf{Wai Keen Vong$^2$}, \textbf{Brenden M. Lake $^{1,2}$} and \textbf{Todd M. Gureckis$^1$}\\
\normalsize $^1$Department of Psychology, $^2$Center for Data Science\\
\normalsize New York University
}

\begin{document}
\maketitle

\begin{abstract}
The Abstraction and Reasoning Corpus (ARC) is a visual program synthesis benchmark designed to test challenging out-of-distribution generalization in humans and machines. Since 2019, limited progress has been observed on the challenge using existing artificial intelligence methods. Comparing human and machine performance is important for the validity of the benchmark.   While previous work explored how well humans can solve tasks from the ARC benchmark, they either did so using only a subset of tasks from the original dataset, or from variants of ARC, and therefore only provided a tentative estimate of human performance. In this work, we obtain a more robust estimate of human performance by evaluating 1729 humans on the full set of 400 training and 400 evaluation tasks from the original ARC problem set. We estimate that average human performance lies between 73.3\% and 77.2\% correct with a reported empirical average of 76.2\% on the training set, and between 55.9\% and 68.9\% correct with a reported empirical average of 64.2\% on the public evaluation set. However, we also find that 790 out of the 800 tasks were solvable by at least one person in three attempts, suggesting that the vast majority of the publicly available ARC tasks are in principle solvable by typical crowd-workers recruited over the internet. Notably, while these numbers are slightly lower than earlier estimates, human performance still greatly exceeds current state-of-the-art approaches for solving ARC. To facilitate research on ARC, we publicly release our dataset, called H-ARC (human-ARC), which includes all of the submissions and action traces from human participants.\footnote{\url{https://arc-visualizations.github.io}}
  
\end{abstract}

\section{Introduction}
In the last several years, large language models (LLMs) have reached impressive performance on a wide variety of benchmarks, demonstrating competency in natural language understanding, coding and mathematics \cite{achiam2023gpt, wei2022emergent}. With larger and more powerful LLMs, many benchmarks have had a limited shelf life, with performance rapidly increasing to human or superhuman levels \cite{kaplan2020scalinglawsneurallanguage}. In contrast, The Abstraction and Reasoning Corpus (ARC, \cite{chollet2019measureintelligence}) has proven to be a persistent and formidable challenge for state-of-the-art AI systems.

\begin{figure}
    \centering
    \includegraphics[width=0.95\textwidth]{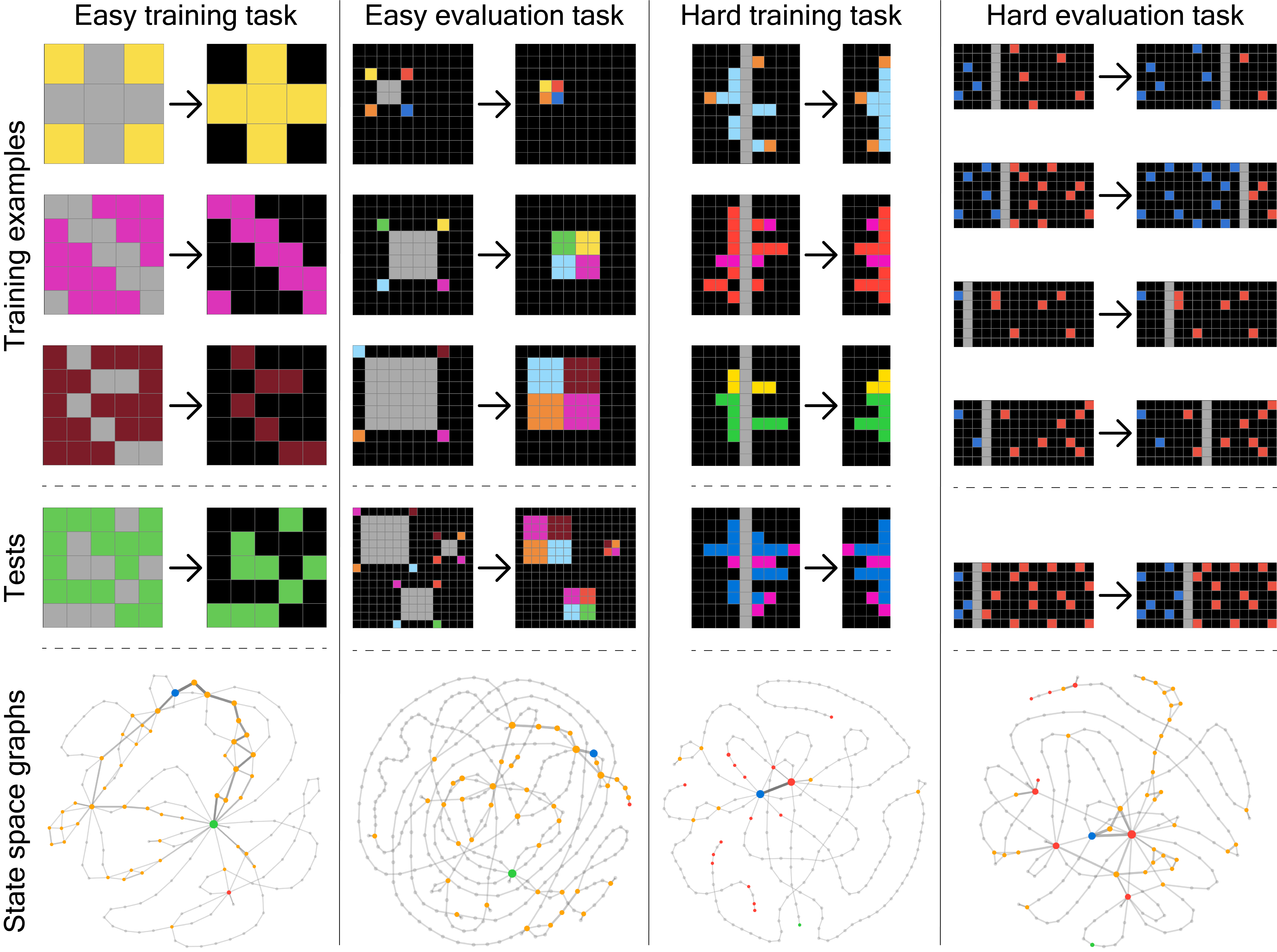}
    \caption{\textbf{ARC Demonstration Tasks}. Shown above are examples of easy (nearly everyone solved them in two attempts or less) and hard (almost no one solved them in two attempts or less) tasks with corresponding training examples and test for each. Interestingly, the easy evaluation task shown here has not been solved by the top LLM solutions to ARC reported in this document. Below the tasks are state space graphs representing all visited states by participants, from starting state (blue nodes) to correct or incorrect submitted grid (green and red nodes respectively). From left to right: \texttt{f76d97a5.json.}, \texttt{e9ac8c9e.json}, \texttt{e3497940.json} and \texttt{dd2401ed.json}.} 
    \label{fig:task-demo}
\end{figure}

The ARC benchmark \cite{chollet2019measureintelligence} was designed to evaluate broad generalization, measuring how algorithms handle a broad category of novel tasks given just a few examples each. Each task requires inferring an underlying transformation rule or program from a series of training input-output pairs which consist of abstract visual grids (see Figure \ref{fig:task-demo}), and to use this rule to correctly generate an output grid given a novel test input. Although visually simple, the tasks are conceptually rich and challenging, requiring the identification of compositional rules involving objects and relations, geometry, counting, visual instructions, and logical operations.

Previous attempts at benchmarking human performance on ARC found human accuracy to be 83.8\% correct, which was estimated using a semi-randomly selected subset of 40 tasks from the training set \cite{Johnson2021-cn}. However, since this result was obtained using only a subset of tasks from the training set, it was unclear how robust this estimate of human performance is, nor whether the same level of performance is achievable on the evaluation set, which is believed to be much harder. The current study aimed to close this gap by providing a robust estimate of human performance on the ARC benchmark, based on human attempts to solve all 400 training and evaluation tasks respectively. We also publicly release H-ARC as the resulting dataset which consists of a total of 15744 attempts on ARC tasks with step-by-step action traces to facilitate further progress for developing more intelligent and human-like systems. Additionally, from the perspective of cognitive science, we believe this an invaluable dataset for enriching our understanding of how people solve a range of novel problems. Although research using variants of ARC tasks \cite{moskvichev2023conceptarc, mitchell2023comparing} and a modified experimental setup \cite{acquaviva2022communicating} have also reported estimates of human performance, to the best of our knowledge, there was no comprehensive estimate of human performance on both the training and evaluation sets prior to this work.

\section{Methods}

\subsection{Design}
We collected human data on each of the 400 training tasks and 400 evaluation tasks from ARC in two separate phases (extending the subset of 40 training tasks previously collected and described in \cite{Johnson2021-cn}). Each task has 1--10 training input-output pairs, and 1--3 test input-output pairs. While only a small proportion of tasks have multiple test input-output pairs (16 and 19 pairs in the training and evaluation set respectively), we opted to evaluate humans using only the first test example for all tasks. On average, 11.8 participants completed each of the 400 training tasks while 10.3 participants completed each of the 400 evaluation tasks.

\subsection{Participants}

We recruited 784 participants (60.2\% male, 37.3\% female, 2.5\% other) on the training set tasks and 948 participants (51.1\% male, 46.1\% female, 2.8\% other) on the evaluation set tasks from Amazon Mechanical Turk using the CloudResearch\footnote{\url
{https://cloudresearch.com}} platform to ensure high quality data \cite{hauser2023evaluating}. Participants were between 18 and 77 years old (\textit{M}=39.8, \textit{SD}=10.4). They were compensated \$10 and were also given a bonus of \$1 if they succeeded at a randomly selected task and its written solution description was judged adequate by the experimenters.\footnote{Best judgement was used: if a description was at least one complete sentence and was relevant to the task, it was counted as adequate.}

\subsection{Experiment}
We evaluated humans using the same evaluation procedure proposed in the original paper describing the ARC benchmark \cite{chollet2019measureintelligence}. In particular, human participants were allowed three attempts per task to generate a correct solution, and were only given minimal feedback on whether each submission attempt was correct or not.

\begin{figure}
    \centering
    \includegraphics[width=0.8\textwidth]{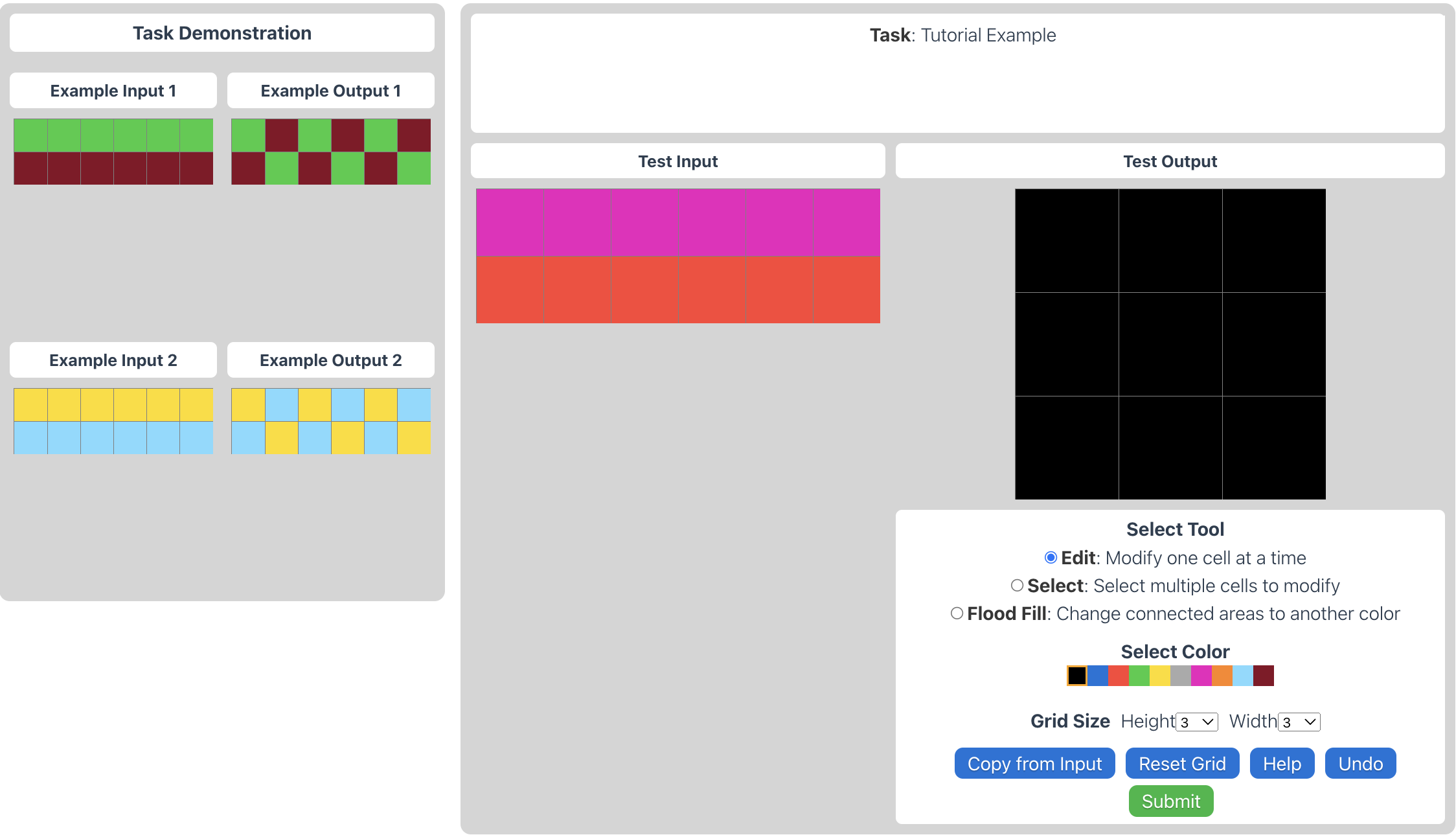}
    \caption{\textbf{ARC Experiment Interface}. Participants were given instructions about the different controls and layout of the interface followed by a tutorial task. Shown here is the evaluation tutorial task \texttt{e9afcf9a.json.}}
    \label{fig:interface}
\end{figure}

\textbf{User Interface}. Participants were first given instructions about the experiment and explanations about the different aspects of the ARC user interface. As in previous experiments \citep{Johnson2021-cn}, the user interface closely mirrored the original interface proposed by \cite{chollet2019measureintelligence} (see Figure \ref{fig:interface}). The interface allowed participants to select different colors and either edit one cell at a time or multiple selected cells. More sophisticated tools allowed the participant to copy and paste a selection from the test input to the test output grid or use the flood fill tool to change the color of all neighbouring cells of the same color to a new color. Participants could resize the grid height and width as well as copy the full test input grid to the test output grid. A reset button allowed participants to revert the output grid back to the initial state, a 3 $\times$ 3 black grid. Finally, unlike in previous iterations of the interface, we added an additional tool allowing participants to undo actions and revert the state of the output grid to the previous state before the last action was taken. At any point in time, the participant could click the help button to display the full set of instructions.

\textbf{Tutorial}: At the beginning of the experiment, participants were provided with animated instructions outlining the user interface with an example task, and then asked to solve the same task to familiarize themselves with the interface. A relatively simple task\footnote{For the training set experiment, we chose task \texttt{21f83797.json} from the evaluation set, whereas for the evaluation set experiment, we chose task \texttt{e9afcf9a.json} from the training set.} was given to participants for the tutorial and they were required to generate the correct test output to proceed (see Figure \ref{fig:interface} for an example). After the tutorial, participants were asked to answer several comprehension questions to make sure they understood the instructions. The experiment started immediately after successful completion of the quiz. Participants were given unlimited attempts at the quiz.

\begin{figure}[h]
    \centering
    \includegraphics[width=0.75\textwidth]{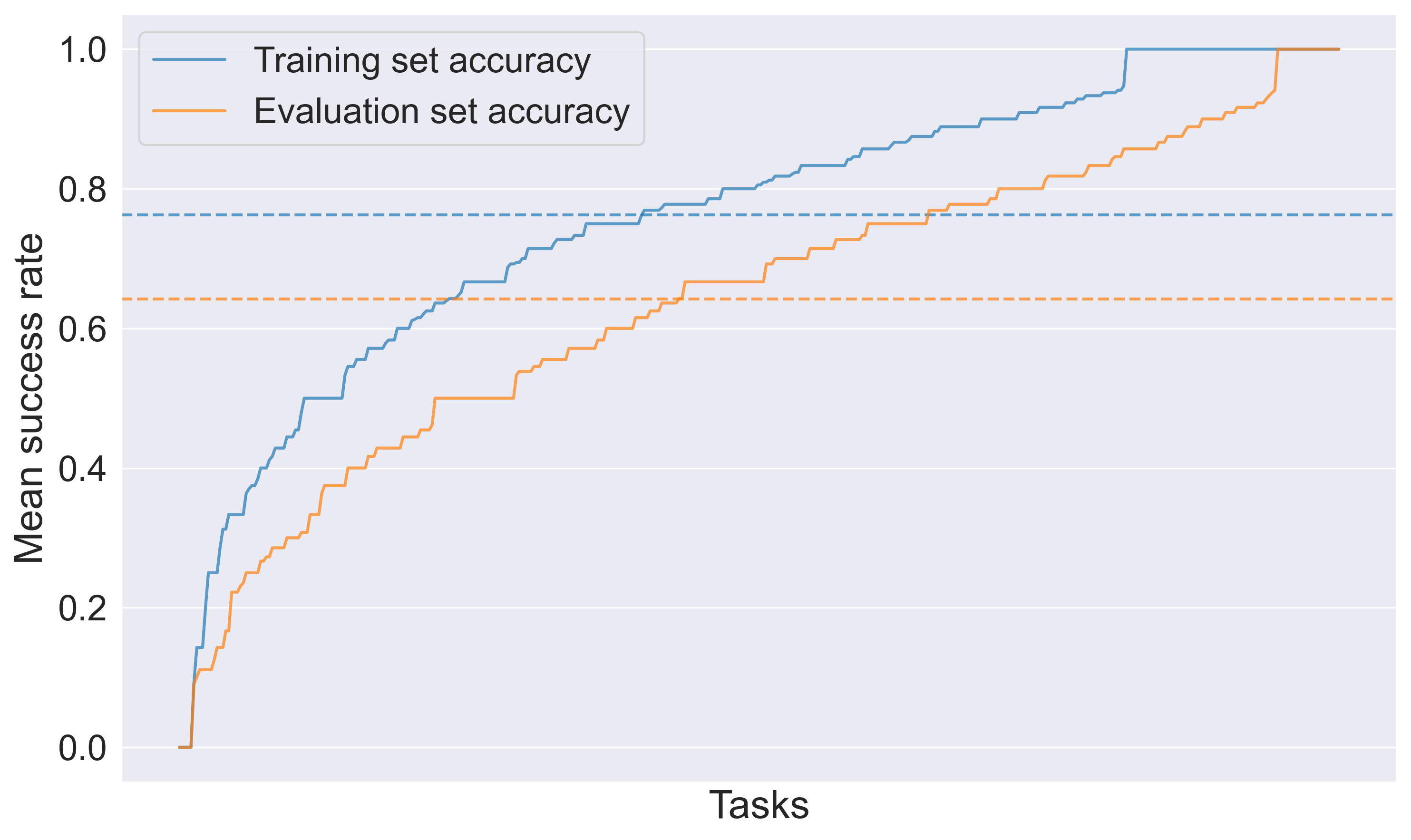}
    \caption{\textbf{Mean 3-shot task success rate.} Tasks are ordered from lowest success rate to highest, showing the distribution of empirically estimated task difficulty for both the 400 tasks in the training and evaluation sets. Dotted lines show average accuracy across all tasks in either the training (blue) or evaluation (orange) split.}
    \label{fig:acc}
\end{figure}

\textbf{Procedure}: The experiment consisted of 5 ARC tasks which were randomly selected from either the set of 400 training tasks or from the 400 evaluation tasks.\footnote{To reduce the potential for attrition or dropouts, we reduced the amount of ARC tasks participants were required to solve from 10 to 5 tasks after collecting 241 out of 783 participants from the first phase of data collection on the training set.} On average, participants completed the experiment in 23 minutes and 1 second (\textit{SD}=13m 24s) on the training set, and 28 minutes and 51 seconds (\textit{SD}=16m 19s) on the evaluation set. There was no time limit for completing a task.\footnote{Participants who exceeded the total time limit of 90 minutes were dealt with manually by email and were included in our dataset nonetheless.} 

For each task, participants were given three attempts. After each attempt, feedback was given on whether the submitted solution was correct or not. Participants were not allowed to resubmit a previously incorrect output grid, ensuring that each of their attempts would be unique.\footnote{We implemented this feature after collecting data from the first 340 participants in the training set experiment. Prior to that, we observed that approximately 8\% of incorrect second and third submission attempts were the same as earlier submission attempts on the same task.} If the participant failed to generate the solution after three attempts, they automatically proceeded onto the next task. 

We also collected natural language descriptions of the inferred solutions by asking participants to write down their solution in words. This was first done after submitting an initial attempt before any feedback was given. If the initial submission was incorrect, participants were asked to submit a second natural language description, either after a subsequent correct submission or on their last (but still incorrect) submission.

\begin{figure}
    \centering
    \includegraphics[width=0.95\textwidth]{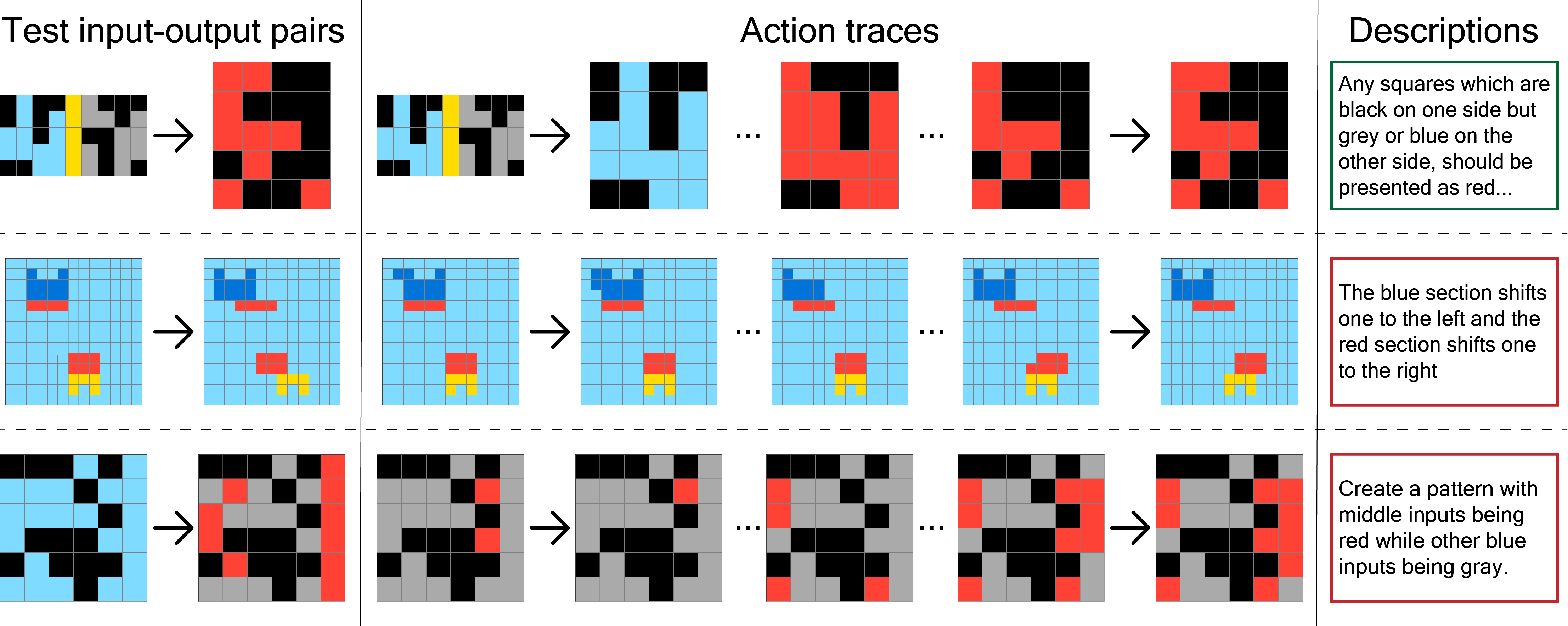}
    \caption{\textbf{Human behavior action traces on ARC problems.} In the left column, we show the test input seen by participants and the true test output grid for three different problems from the ARC evaluation set. In the middle column, sampled action traces show successive states of the grid with the last state corresponding to a correct (green box) or incorrect (red box) submission. In the last column, we show the first natural language descriptions submitted by participants along with their solution. From top to bottom: \texttt{34b99a2b.json}, \texttt{4364c1c4.json} and \texttt{a8610ef7.json}.}
    \label{fig:action-traces}
\end{figure}

\textbf{Incomplete participant data}:
Participant data collected online can be incomplete for many reasons: participants may find the task too hard, have technical difficulties, find the experiment uninteresting, not understand the instructions or even run out of time. In our experiments, a number of participants withdrew from the experiment after completing between 0 and 4 tasks, although most did not provide any particular reason for withdrawing from the experiment. We found that 90 out of 783 participants' data from the training set experiment is incomplete, while 242 out of 946 participants' data from the evaluation set experiment is incomplete. Our results indicate that 7.5\% and 13.3\% of the training and evaluation set task data respectively are missing for a total of 10.3\% missing task data. We obtained these values by computing the proportion of expected task data (number of participants $\times$ number of tasks assigned) that was missing from our dataset. To account for these substantial rates of incomplete data, we report pessimistic and optimistic performance estimates to provide a range of values within which human performance on ARC is likely to lie. This was accomplished by simulating imputed scores to the missing data from participants who dropped out.

\begin{table}
\centering
\begin{tabular}{lcccccc}
\toprule
\multirow{2}{*}{\textbf{Model}} & \multicolumn{3}{c}{\textbf{Training Set} (\%)} & \multicolumn{3}{c}{\textbf{Evaluation Set} (\%)} \\
\cmidrule(lr){2-4} \cmidrule(lr){5-7}
& avg / best & pess & opt & avg / best & pess & opt \\
\midrule
Human 1-shot & 59.9 / 96.8 & 57.6 & 61.5 & 47.8 / 95.8 & 41.6 & 54.6 \\
Human 2-shot & 72.6 / 98.5 & 69.7 & 73.7 & 60.2 / 97.8 & 52.4 & 65.4 \\
Human 3-shot & \textbf{76.2} / \textbf{98.8} & 73.3 & 77.2 & \textbf{64.2} / \textbf{98.8} & 55.9 & 68.9 \\
\midrule
Claude-3.5-\textit{N} (1 / 2-shot) & \multicolumn{3}{c}{-} & \multicolumn{3}{c}{19.3 / 20.7} \\
GPT-4o-\textit{NS} (1 / 2-shot) & \multicolumn{3}{c}{-} & \multicolumn{3}{c}{38.5 / 42.0} \\
\bottomrule
\end{tabular}
\vspace{0.5em}
\caption{\textbf{Human ARC Performance Summary}: Human average (avg) represents empirically estimated human performance after one, two or three attempts, averaged across all tasks in the training and evaluation sets respectively. Human best refers to the overall proportion of tasks that any human participant successfully solved, across all tasks in either the training and evaluation sets. We also report pessimistic (pess) and optimistic (opt) estimates of human performance. For comparison, we also report performance from the current public leaderboard for the ARC challenge. Note that Claude-3.5-\textit{N} was only evaluated on a subset of 150 evaluation tasks.\protect\footnotemark}
\label{tab:ARC-performance}
\end{table}

\begin{table}[h]
\centering
\begin{tabular}{lcc}
\toprule
\textbf{Metric} & \textbf{Training Set} & \textbf{Evaluation Set} \\ 
\midrule
Number of tasks & 400 & 400 \\
Total number of participants & 783 & 946 \\
Incomplete participants & 90 & 242\\  
Average participants per task & 11.8 & 10.3 \\
Average attempts to solution & 1.3 & 1.4 \\
Total attempts & 7924 & 7820\\
Unique number of visited states & 127146 & 208214\\
Total action traces & 241697 & 344569\\
\bottomrule
\end{tabular}
\vspace{0.5em}
\caption{\textbf{Human ARC Descriptives}: Here we report numerical values summarizing our behavioral dataset. ``Total attempts'' represents the number of individual submissions across all tasks and ``total action traces'' represents the number of single actions recorded across all tasks and participants.}
\label{tab:ARC-descriptives}
\end{table}

\footnotetext{\url{https://arcprize.org/leaderboard}}

\section{Results}

\subsection{Performance}
\begin{figure}[h]
    \centering
    \includegraphics[width=0.75\textwidth]{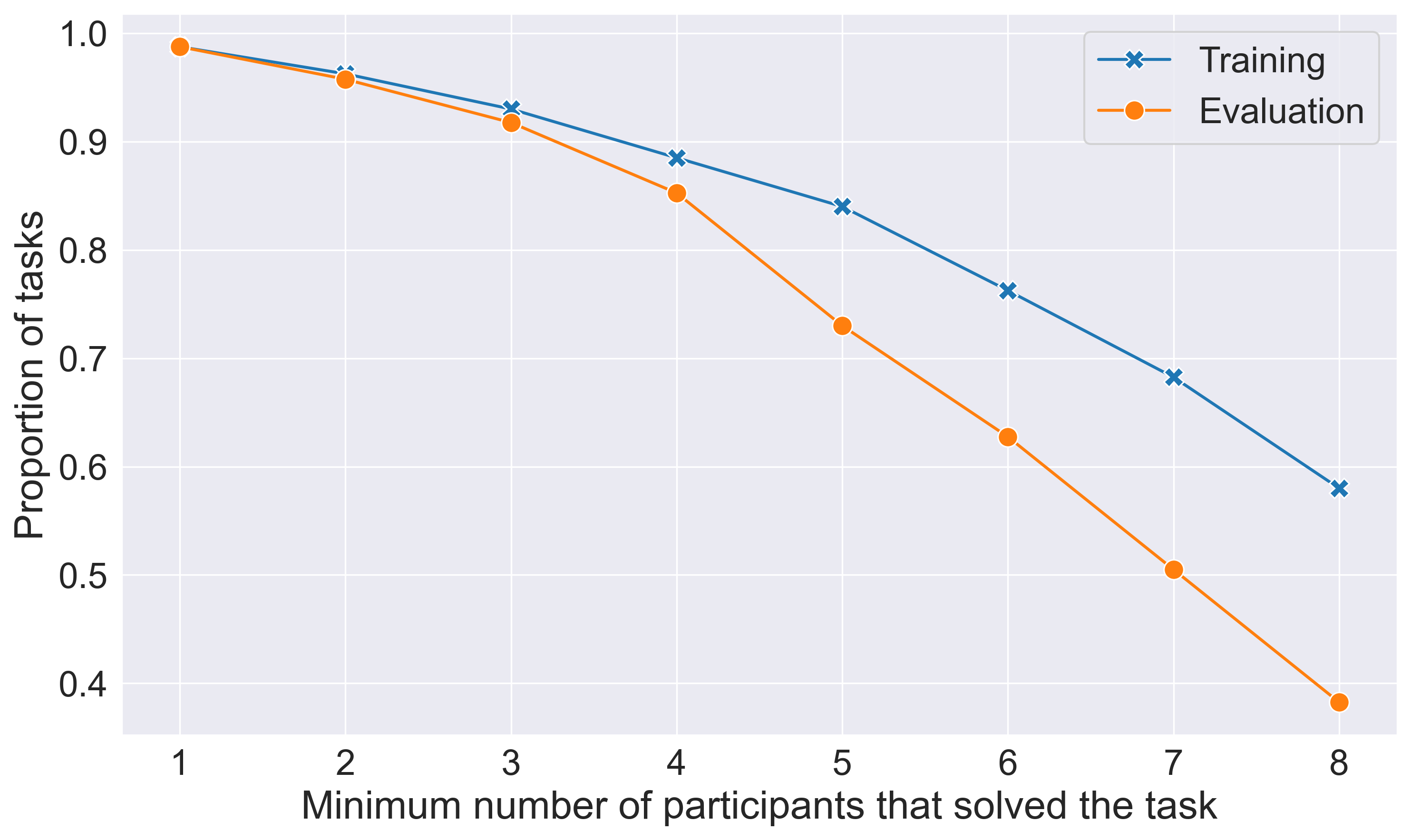}
    \caption{\textbf{Minimum number of correct submissions across tasks.} In this figure, we report the proportion of tasks where a minimum of N (x-axis) participants submitted a correct solution in three attempts or less. Note that higher values of N are biased negatively since posterior inference on a binomial outcome with small sample size is skewed.}
    \label{fig:min_success}
\end{figure}

For both ARC datasets, we report a range of performance values that reflect different ways of thinking about our estimate of human performance on ARC tasks and its inherent uncertainty (see Figure \ref{fig:min_success}). In general, we find that human performance is higher on the training set compared to the evaluation set of ARC, validating previous intuitions about the relative difficulty between the two splits. Yet, despite the drop in performance on the evaluation set, humans still greatly outperform current state-of-the-art approaches to ARC, and we report and compare performance with two recent LLM-based solutions from the ARC prize public leaderboard (see Table \ref{tab:ARC-performance}). 

The first LLM-based solution \cite{kamradt2024frontier} we report involves minimal prompting of Claude 3.5 Sonnet \cite{Anthropic2024-zl} using text-only representations of ARC problems. Input and output training examples as well as test inputs are labelled and represented as lists of lists. The model is instructed to generate a JSON formatted list of lists response representing the output grid after applying the inferred pattern to the test input. This model was only evaluated on a subset of 150 tasks from the evaluation set. Throughout the rest of this report, we will simply refer to this approach as ``Claude-3.5-\textit{N}'' for Claude-3.5-\textit{neural}. The second LLM-based solution \cite{greenblatt2024} we report involves elaborate few-shot prompting of GPT-4o \cite{achiam2023gpt} with both images and text representations of ARC problems as well as instructions about how to reason about the task. The model is instructed to generate code that parses the test input to produce an output grid and also attempts to make corrections to its own generated code. After extensive sampling of the model (approximately 8000 code completions), the top 3 outputs are submitted using a majority vote over programs. We will refer to this method as ``GPT-4o-\textit{NS}'' for GPT-4o-\textit{neurosymbolic} since it explicitly combines a neural and symbolic, code-based approach. 

\textbf{Training set.} We computed accuracy by calculating the proportion of successful submissions on each task after three attempts or less and averaged across all tasks. We also estimate pessimistic and optimistic performance values. In the pessimistic case, for each participant that completed $k < 5$ tasks, we sample $5-k$ random tasks and assume failure. Conversely, in the optimistic case, we repeat the same procedure but assume success on every sampled task.\footnote{We consider these pessimistic and optimistic estimates somewhat unrealistic. In the pessimistic case, there are many reasons a participant might have dropped out of the task, including running out of time, technical issues, or lack of interest.  There is no reason to assume that \textit{all} remaining tasks would have been failed.  Similarly, in the optimistic case there is no reason to think that people would succeed on every subsequent task they did not attempt. However, these bounds help us bracket how much the drop out effect might have altered our estimates.} We run 1000 simulations for each case, imputing missing data using the sampled tasks and outcomes. We then take the resulting average mean task success rate to compute pessimistic and optimistic estimates which we report in brackets. Our results suggest an estimated average task accuracy of 76.2\% (\textit{SD}=21.5\%, [73.3\%, 77.2\%]) on the training set of ARC. We also report average task accuracy based on participants' first and second attempts, resulting in average task accuracy of 59.9\% (\textit{SD}=24.8\%, [57.6\%, 61.5\%]) and 72.6\% (\textit{SD}=22.9\%, [69.7\%, 73.7\%]) respectively.

\begin{figure}
    \centering
    \includegraphics[width=0.75\textwidth]{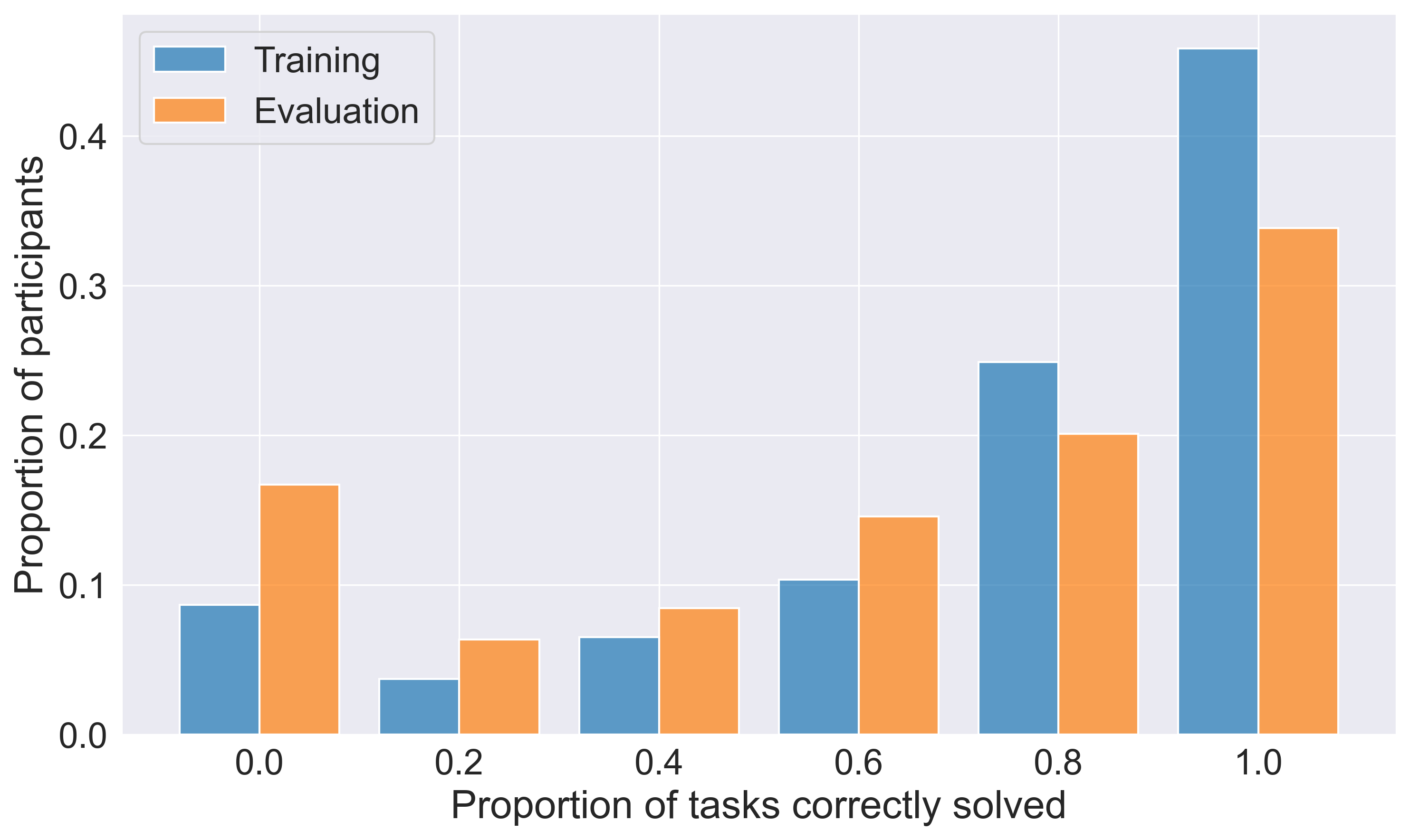}
    \caption{\textbf{Proportion of tasks solved by participants.} We report the proportion of tasks solved instead of discrete number of tasks since there were a minority of participants that completed a different number of tasks (e.g. due to dropping out, or having completed an earlier version of the experiment with ten distinct tasks).}
    \label{fig:success_dist}
\end{figure}

Participants solve ARC training tasks in 1.3 attempts on average, with the modal and median number of attempts being one. Of the 400 training tasks, we find 74 tasks (18.5\% of the training set) for which all participant who attempted the task generated the correct solution after three submissions or less. Conversely, we also find 5 tasks (1.3\% of the training set) which no participants were able to solve correctly after three attempts (see Figure \ref{fig:acc}).\footnote{Note that since each problem is attempted by approximately 10 people, this result simply means that we did not find anyone in a set of 10 that could solve the problem, not that they are in-principle not solvable.} Finally, we find that 40.1\% of participants solved all training set tasks they were presented and that 8.6\% of participants solved none (see Figure \ref{fig:success_dist}).

\textbf{Evaluation set.} Independent samples t-tests suggest that evaluation tasks are significantly harder for people than training tasks, $t(798)=7.67, p<.001$. We estimate that the average task accuracy after three attempts on the evaluation set is 64.2\% (\textit{SD}=22.8\%, [55.9\%, 68.9\%]). In addition to this result, we report a first and second attempt average task accuracy of 47.8\% (\textit{SD}=23.2\%, [41.6\%, 54.6\%] and 60.2\% (\textit{SD}=23.3\%, [52.4\%, 65.4\%]) respectively.

\begin{figure}[h]
    \centering
    \includegraphics[width=0.75\textwidth]{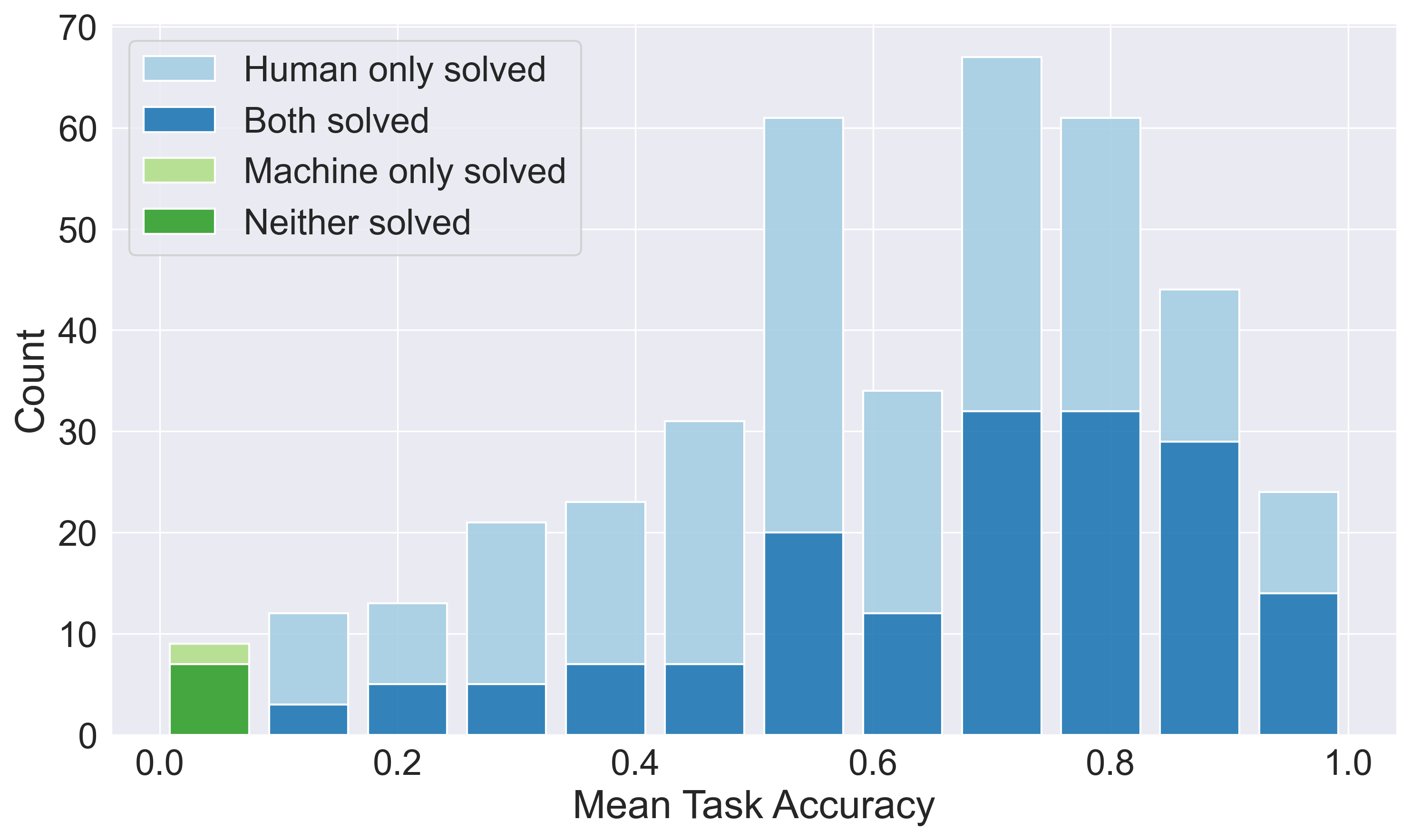}
    \caption{\textbf{Human-machine performance comparison.} In this figure, we compare human (2-shot) performance with GPT-4o-\textit{NS}. We show the proportion of tasks solved by humans or machine only, both and neither for varying levels of human performance. We find 225 tasks that were only solved by humans, 7 tasks solved by neither and 2 tasks solved only by GPT-4o-\textit{NS}.}
    \label{fig:human-machine}
\end{figure}

On average participants solve ARC evaluation tasks in 1.4 attempts, with the modal and median number of attempts being one. Of the 400 evaluation tasks, we find 22 tasks (5.5\% of the evaluation set) for which participants always found the correct solution. Conversely, we find 5 tasks (1.3\% of the evaluation set) which no participants were able to solve (see Figure \ref{fig:acc}). We also find that 33.8\% of participants solved all evaluation set tasks they tried and that 16.7\% solved none (see Figure \ref{fig:success_dist}). After comparing 2-shot success across evaluation set tasks for people and GPT-4o-\textit{NS}, we find 225 tasks that were only solved by humans, 7 tasks solved by neither and 2 tasks solved only by GPT-4o-\textit{NS} (see Figure \ref{fig:human-machine}).

\subsection{Cross-dataset comparisons}
Although we don't find any conclusive indicators of why the evaluation set is harder than the training set, we report our attempt to understand what drives the increased difficulty. In particular, we examined two aspects of each dataset which we hypothesized to be contributors to difficulty: test output grid size and time spent on problems.

\textbf{Output grid size.} An independent samples t-test confirmed that output grid size (number of grid cells) is significantly larger on average in the evaluation set (\textit{M}=235.1, \textit{SD}=246.7) than in the training set (\textit{M}=136.2, \textit{SD}=164.9), $t(798)=6.81, p<.001$. Although we find that grid size is mildly correlated with difficulty on the training set, $r(398)=-0.16, p=.001$, we did not find a similar relationship between output grid size and performance in the evaluation set $r(398)=-0.02, p=.649$.

% \begin{figure}
%     \centering
%     \includegraphics[width=0.75\textwidth]{figures/mean_task_accuracy_vs_grid_size.png}
%     \caption{\textbf{Task accuracy as a function of grid size.}}
%     \label{fig:success_dist}
% \end{figure}

\textbf{Time spent on problems.}
Reaction time or time spent thinking about a problem, has been shown to reflect value of computation in chess games \cite{russek_2022_time}, relates effort allocation and resulting scores in IQ tests \cite{cheyette_2024_iq} and has a strong history in psychology as a window into underlying cognitive processes. We performed a number of temporal analyses based on various timing-related aspects of the human behavioral data.

% Following this line of research, we report preliminary results from our temporal analyses.

First, we computed the time taken from seeing a new task to submitting a final solution for each participant and task. We then normalized these values by the number of attempts taken by each participant on each task. An independent samples t-test showed that participants spend significantly less time solving tasks from the training set (\textit{M}=4m 19s, \textit{SD}=4m 4s) than tasks from the evaluation set (\textit{M}=6m 39s, \textit{SD}=5m 59s), $t(8874)=-21.63, p<.001$. Surprisingly, average performance on the evaluation set was lower despite participants spending more time and effort on average per task compared to the training tasks.  

Second, how much time do people spend thinking instead of acting? As a first approximation of how much time is spent \textit{thinking} about ARC puzzles, for each attempt per participant, we computed the sequence of inter-action times: the time spent between each action (i.e., clicks) taken to modify the state of the output grid or otherwise interact with the user interface. We then filtered out any inter-action time interval smaller than 5 seconds, corrected for time spent writing and only considered series of submissions that led to a correct solution. Finally, we sum the remaining inter-action time intervals and normalized by the number of attempts taken to solve each task. An independent samples t-test confirmed that participants spend significantly less time thinking about training set tasks (\textit{M}=1m 6s, \textit{SD}=1m 10s) than they do about evaluation set tasks (\textit{M}=1m 37s, \textit{SD}=1m 39s), $t(1505)=-6.90, p<.001$. This result provides evidence that more thinking is required to infer the underlying rule or program of evaluation tasks as opposed to training tasks, and suggests that the reasoning process underlying finding a solution for evaluation tasks is more computationally demanding than for training tasks. 

% \subsection{Action traces}
% Longer action traces? Mo  re divergent action traces? More / less bottleneck states (if time?)

% The data generated by participants results in rich step-by-step action traces that could be revealing of cognitive processes underlying solving ARC tasks. We find that people use tools in such and such way.

% \subsection{Natural language}

\subsection{Errors}

Another aspect of the behavioral data that is potentially revealing of the underlying cognitive processes involved in solving ARC problems is looking at the types and patterns of errors that participants make (see Figure \ref{fig:action-traces}). We explore a number of preliminary analyses looking at errors along a variety of dimensions: errors on the dimensions of the test output grid, edit distance from the true test output, error divergence and copying, and compare them to the machine outputs reported in this document (see Table \ref{tab:ARC-performance}).

\textbf{Grid dimension errors.} 
One of the most common first actions that people take on both training and evaluation set tasks (33.2\%) is to resize the output grid to the intended size by selecting the height or width drop-down menu. Although people make a non-negligible amount of height and width errors, within the set of incorrect submissions, we find that 68.2\% and 73.5\% of submission attempts have both the correct height and width in training and evaluation set tasks respectively. Conversely, we find that both current top solutions to the public ARC leaderboard reported in this document (see Table \ref{tab:ARC-performance}) make fewer mistakes in selecting the correct height and width of the test output compared to humans, with Claude-3.5-\textit{N} making grid dimension errors on 10.3\% of incorrect submissions and GPT-4o-\textit{NS} making grid dimension errors 8.3\% of the time.

\textbf{Edit distance.} As a proxy for how close to the true test output incorrect submissions tend to be for ARC tasks, we computed the distribution of edit distances for all errors on each task.\footnote{Edit distance may be an imperfect proxy for errors, as large edit distances may still arise from inferring an almost correct rule or program to a given task.} To facilitate comparison, we computed edit distance on grids with correct dimensions and normalized by grid size. We find that edit distance to ground truth distributions for GPT-4o-\textit{NS} (\textit{M}=0.19, \textit{SD}=0.18), Claude-3.5-\textit{N} (\textit{M}=0.19, \textit{SD}=0.16) and humans (\textit{M}=0.19, \textit{SD}=0.15) are strikingly similar. Nonetheless, we find an average normalized pairwise distance between human errors and machine errors across tasks of 0.25 for both reported models ($SD_{GPT-4o}$=0.19, $SD_{Claude3.5}$=0.20). Despite people and machines having similar edit distance to ground truth distributions, this result clearly suggests that they are making substantially different errors.

\textbf{Error divergence}. We computed error divergence as the number of unique incorrect grids over the total number of submitted grids for each problem. Task success rate and error divergence were found to be strongly negatively correlated on training tasks, $r(378)=-0.81, p<.001$, and on evaluation tasks $r(395)=-0.88, p<.001$. Since we do not have repeated samples data from models, we omit comparative analyses here.

\textbf{Copy errors}. We also find that 10.0\% and 9.2\% of errors on training and evaluation set tasks respectively were copy errors. Copy errors refer to any incorrect submission that was a copy of either the example inputs (0.6\% and 0.3\% for training and evaluation sets respectively), example outputs (2.4\% and 3.6\% for training and evaluation sets respectively) or the test input (7.0\% and 5.4\% for training and evaluation sets respectively). In contrast, only 3.5\% and 1.6\% of incorrect outputs from GPT-4o-\textit{NS} and Claude-3.5-\textit{N} were copy errors, respectively.

\section{Discussion}
The present study aimed to provide a comprehensive evaluation of human performance on the ARC benchmark. To estimate performance on ARC, we collected data from 1729 Amazon Mechanical Turk workers who were each assigned 5 randomly selected tasks from either the training or evaluation set. We estimate that 3-shot human performance is between 73.3\% and 77.2\% on the training set with an observed empirical average of 76.2\%, and between 55.9\% and 68.9\% on the evaluation set with an observed empirical average of 64.2\%. We also report that 98.8\% of both the training and evaluation sets are solved by at least one person. Overall, we find that the evaluation set is more difficult for people than the training set. Although it remains unclear why evaluation set ARC puzzles are harder, we find that people spend significantly more time thinking about evaluation set tasks than they do about training tasks. Finally, we analyzed error patterns related to grid dimension, edit distance, error divergence and copying for both people and machines. Although people solve more tasks than state-of-the-art approaches, we find that machines outperform people on most error metrics we analyzed. These results along with edit distance results suggest that the errors people and machines make on ARC tasks are of a different nature, further emphasizing that these approaches do not capture how people solve ARC problems. We discuss these issues further in the following sections.

% we fit a hierarchical IRT model to analyze participant performance on both the training and evaluation sets. Our model accounts for and estimates latent participant ability, task difficulty and the effect of number of attempts on success on a task. Our results show that the estimated performance on ARC training tasks is 

\subsection{Competence versus performance: contextualizing average performance}
It is important to bear in mind that average performance on ARC tasks as reported here can be affected by many contextual factors and does not reflect some absolute measure of human performance. For example, if participants were paid more for only correct responses, were put under more extreme time pressure, or if participants were sampled from a broader population we might find different values of average performance. The average scores simply reflect ``how a paid Internet worker from Amazon Mechanical Turk will typically do at the given pay rate" rather than a universal estimate of human ability. For this reason the performance measures might have less durability than measures of human \textit{competence} (i.e., can any human, in principle, solve the problems?). That said, the estimates we provide here still exceed those of state-of-the-art AI algorithms applied to the task, suggesting a substantial human-machine intelligence gap.

\subsection{Human competence}
For almost every task (98.8\%) in the combined ARC training and evaluation sets, there is at least one person that solved it and over 90\% of tasks were solved by at least three randomly assigned online participants. This is interesting because effectively this means that if you contact 10 random people on the Internet (via CloudResearch), at least one will be able to solve any given ARC problem. This is in contrast to biased sampling from a selective group like AI researchers (including the inventor of the ARC problem set) or highly educated academics. And this estimate is clearly biased to be low because only 10 people completed each problem with three attempts (if we had 1000 people complete each problem, or gave each person unlimited attempts, we would expect the odds of a human solving every problem would go up).  The point is we didn't have to do this exhaustive sampling to find almost universal solvability of the ARC problems.  We believe this critically highlights that human intelligence is \textit{in principle} capable of carrying out the required computations to solve any ARC task which is unlike any AI model reported so far. This result would appear to put aside doubts about the solvability of ARC tasks by humans in both the training and evaluation sets. Furthermore, similarly to how expertise at games like chess and Go had been pitted as the threshold for successful AI models, our human best performance score suggests a challenging goal for AI.

\subsection{Self-correction through minimal feedback}
Although people make errors, our analyses as well as qualitative judgements suggest that people are better at learning from minimal feedback, and correcting for those errors than machines. In fact, most correct answers from either top solution reported here are obtained on a first attempt, with only +7\% and +9.1\% proportional increase in accuracy with a second attempt for Claude-3.5-\textit{N} and GPT-4o-\textit{NS} respectively. For humans, we see substantial improvements in accuracy after 2 attempts (+21.2\% on the training set and +25.9\% on the evaluation set) and still more proportional increase in accuracy when given 3 attempts (+5.0\% on the training set and +10.0\% on the evaluation set). People will often make initially wrong guesses but they are capable of self-correction and can flexibly consider alternative solutions. Understanding how people achieve this is likely to be useful for improving machine intelligence in ARC tasks, and more generally for problem-solving.

\subsection{Why is the evaluation set more difficult?}
The evaluation set is found to be significantly more difficult than the training set for people. Although it is still unclear why that is the case, our results suggest that factors other than output grid size are contributing to difficulty on the evaluation set and that for equally sized output grids, evaluation tasks are still often more difficult than training tasks. We suspect that the primitive operations underlying the transformations for evaluation tasks are more difficult to infer and/or execute than in the training set. For instance, we have found that logic-based operations that require superimposing grid sections tend to be difficult for people (see Figure \ref{fig:action-traces}). 

ARC problems present an interesting challenge: how can the outputs of a program given its inputs be decomposed into subroutines and primitive operations that recover the underlying program? Previous work has explored a hypothesis-generation model using an LLM to solve a subset of ARC tasks \cite{wang2023hypothesis}. Results from this research demonstrated that an important bottleneck on LLM performance in this setting was generating accurate and useful natural language hypotheses about the underlying program or rule of each task. Another line of research has found that nameability or codability can modulate concept acquisition \cite{Zettersten2020-nm}, making it harder to learn concepts that have features that are more difficult to name. Similarly, we believe that certain primitive operations underlying ARC tasks in the evaluation set might be more difficult to identify or retrieve for people, in turn making it more difficult to generate plausible hypotheses about the underlying program. Another possible source of difficulty is that the complexity or description length of programs in the evaluation set may be 
longer on average than in the training set. We hope that the natural language descriptions and action traces from H-ARC will help elucidate these questions.

\section{Conclusion}

Although we provide a comprehensive estimate of human performance on ARC tasks and preliminary analyses, more in-depth analyses of essential elements of people's problem solving strategies such as state space trajectories, action traces and natural language descriptions are needed. H-ARC affords answering questions about these aspects of behavior which are likely to be informative for understanding the underlying mental representations that support abstract reasoning and problem-solving in people. As a result, we hope that the dataset we release along with this document will allow developments on two fronts: our understanding of how people solve novel abstract problems and how machines can be improved to reason about such problems in a more human-like and intelligent way.

\subsubsection*{Acknowledgements}
We thank Mark Ho for comments and feedback during the revision of this document, Aysja Johnson for collecting the data from the first few hundred participants on the training set and Nicholas Emery Kirsch for funding the data collection and personnel efforts on the evaluation set. This work was also generously supported by NSF BCS grant 2121102 to TMG.

\bibliographystyle{apalike} 
\bibliography{human-arc}
\end{document}